\documentclass[preprint,review,12pt]{elsarticle}

\usepackage{amssymb}
\usepackage{amsmath}

\usepackage{multirow}              
\usepackage{booktabs}               
\usepackage{setspace}               
\usepackage{lineno}                 
\usepackage{array}                   
\usepackage{url}                     
\usepackage{amsfonts}                

\begin{document}
	
	\begin{frontmatter}
		
		
		
		\title{Phase-Aware Spatial-Frequency Fusion for Few-Shot Fine-Grained Image Classification}
		
		
		\author[1]{Ruiling Liu}
		
		\author[1]{Linyue Zhang}
		
		\author[1]{Wenyi Zeng}
		
		\author[1]{Jiamiao Lu}
		
		\author[1]{Weichuang Zhang\corref{cor1}}
		\ead{zwc2003@163.com}
		
		\author[2]{Changming Sun}
		
		\author[3]{Zejun Zhang}
		
		\author[1]{Xiao Zhao\corref{cor1}}
		\ead{zwc2003@163.com}
		
%
%
		
		\cortext[cor1]{Corresponding authors: Weichuang Zhang and Xiao Zhao}
		
		\begin{abstract}
			Few-shot fine-grained image classification (FSFGIC) aims to classify similar images with limited labeled examples. This work highlights the critical yet underutilized role of phase information in capturing structural relationships within an image. This study introduces a novel plug-and-play amplitude-phase integration (API) module that effectively combines local and global frequency amplitude and phase information for obtaining more comprehensive feature descriptors. Additionally, a dedicated network, named PSF-Net, is proposed that adaptively fuses phase-based spatial and frequency information for FSFGIS. The designed PSF-Net can be easily integrated into standard episodic training architectures for end-to-end training from scratch. Extensive experiments on five public datasets demonstrate that the method outperforms existing state-of-the-art benchmarks.
		\end{abstract}
		
		
		
		
	\end{frontmatter}
	
		
		\section{Introduction}
		\label{sec:intro}
		Few-shot fine-grained image classification (FSFGIC)~\cite{ren2025adaptive, wang2026dual,wang2026frequency,wang2026mssffe, zhang2026adaptive} aims to distinguish different yet similar subordinate classes that belong to a specific basic class with very limited data. The existing FSFGIC methods can be roughly classified into two main streams:~meta-learning based methods and metric-learning based methods. Meta-learning based methods aim to learn meta knowledge with only a few training examples for each category. A set of functions are used for mapping labeled training image samples and test samples for visual classification. Metric-learning based methods intend to learn a group of functions for transforming test samples into an embedding space. Then the test samples will be classified into a class by a given similarity measure (e.g., nearest neighbor, hyperbolic distance, cosine metric, or learned parametric options).

\begin{figure}[htbp]
			\centering
			\includegraphics[width=0.7\linewidth]{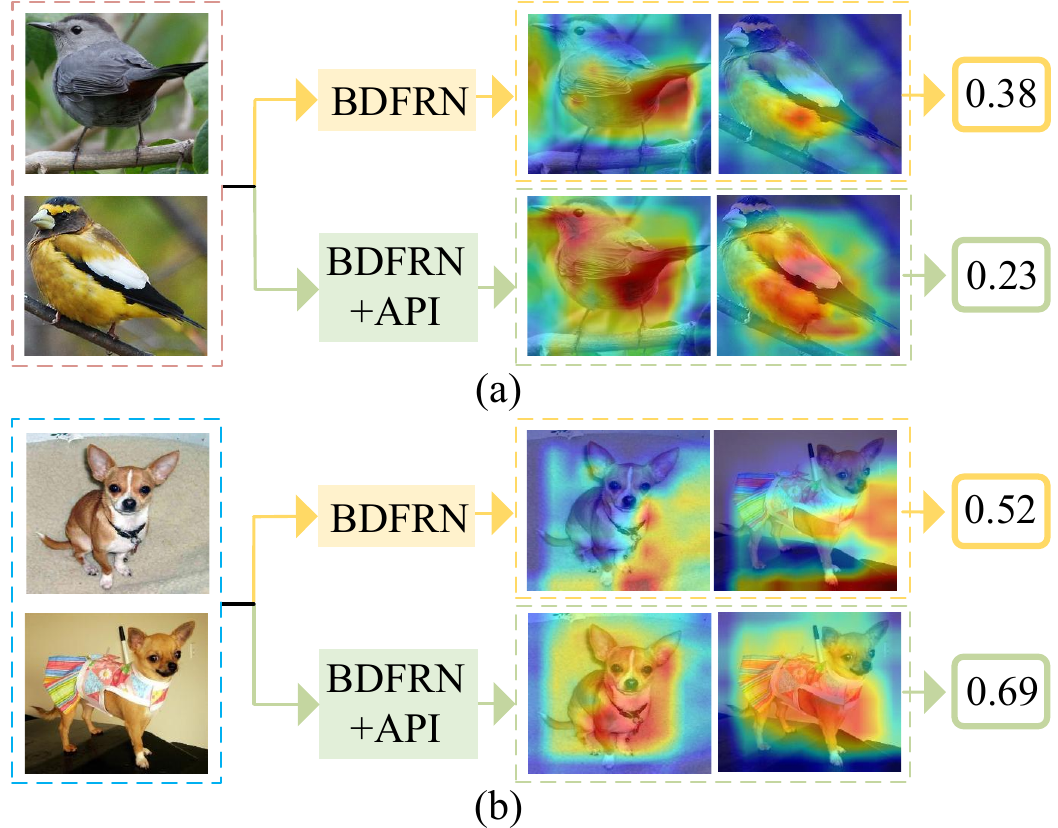}
			\setlength{\abovecaptionskip}{1pt}  
			\setlength{\belowcaptionskip}{-8pt}  
			\caption{The impact of phase information on the FSFGIC accuracy of a given network. }
			\label{fig1}
		\end{figure}

Existing image feature extraction techniques~\cite{wang2026aagdd, jing2022recent,  jing2022image, liu2024aekan, zhang2024re,  qiu2021recurrent,  pan2024pseudo, ren2024few, jing2023ecfrnet,  pan2025overcoming,  lei2024semi, yuan2026gloresnet, pan2024dycr} have shifted from manual algorithms~\cite{shui2013corner, lu2026second, zhang2014corner, zhang2019corner, shui2012noise, zhang2017noise, zhang2015contour, zhang2020corner, gao2020fast, li2019multi, zhang2019discrete,  wang2020corner, li2023traffic, zhang2023image,  xie2026second, bao2022corner, an2023edge, li2023mutual} to deep learning architectures~\cite{ zhang2021ndpnet, ma2023ct, liao2025dynamic,  li2023m, wang2024unbiased, lu2022image, jing2021novel, zheng2023fully,  ren2025adaptive, wang2025principal, liao2022asrsnet, li2024automotive,  liao2023feature, tang2025cascading, ren2025zero, wang2025feature} for feature extraction. Currently, FSFGIC presents significant challenges~\cite{song2025efficient, lu2026meningioma, guo2026gattenrnn, ren2026deep, liao2026neuron, liao2025learning,  zhang2026adaptive, wang2026dual,sun2026highly} due to subtle inter-class differences and substantial intra-class variations within very limited training samples. Some methods~\cite{Cheng2023FGFL} that integrate spatial and frequency domain information have been proposed for addressing these issues and have demonstrated certain improvements. However, our analysis reveals that existing spatial-frequency fusion-based FSFGIC methods may still yield inaccurate classifications, primarily because they tend to overlook the critical role of phase information in the frequency domain. Phase information, which records the positional relationships of different frequency components, is essential for characterizing the structural layout of an image. Neglecting it can lead to a loss of important discriminative cues necessary for distinguishing highly similar sub-categories. Take four fine-grained images as examples as shown in Fig.~\ref{fig1}. The BDFRNet~\cite{wu2024bi1} incorrectly classifies images from different categories as the same class as shown in Fig.~\ref{fig1}(a) and misclassifies images from the same category as different classes as shown in Fig.~\ref{fig1}(b) due to its failure to account for phase information((a) shows different classes samples, (b) shows the same class samples).
		To address the aforementioned problems, in this work, a novel plug-and-play amplitude-phase integration (API) module(Fig.~\ref{fig1}(a)) is presented that has considered how to effectively combine local and global frequency amplitude and phase information for obtaining more comprehensive feature descriptors. Fig.~\ref{fig1}(b) and (c) demonstrate that when equipped with our proposed API module, BDFRNet~\cite{wu2024bi1} has the capability to both cluster similar targets into the same category and differentiate between distinct categories. Furthermore, a dedicated network, named as PSF-Net, is proposed which has considered how to fuse phase information-based spatial and frequency information adaptively for FSFGIS. The designed PSF-Net can be easily integrated into standard episodic training architectures for end-to-end training from scratch. Experiments on five fine-gained image classification benchmark datasets, i.e., CUB-200-2011~\cite{wah2011caltech}, Stanford Dogs~\cite{khosla2011novel}, Stanford Cars~\cite{krause20133d}, meta-iNat~\cite{wertheimer2019few}, and tiered meta-iNat~\cite{wertheimer2019few}, show that the proposed PSF-Net significantly outperforms the baseline methods on both 5-way 1-shot and 5-way 5-shot tasks.
		\section{Related Work}
		\subsection{FSFGIC Methods}
		Meta learning-based FSFGIC methods rapidly acquire model initializations or parameter updates by leveraging three principal techniques: attention mechanisms, feature alignment, and knowledge distillation.Attention-based methods employ mechanisms such as multi-attention~\cite{zhu2020multi}, dual-attention~\cite{Shulin2022}, and transformer-based architecture~\cite{Zhang2023} for amplifying subtle differences between highly similar categories. Feature alignment-based methods~\cite{wei2019piecewise} aim to spatially align corresponding object parts between support and query images to capture fine-grained differences. Techniques include bilinear CNNs and distribution alignment for effective comparison. Knowledge distillation-based methods~\cite{wu2023few} aim to improve model efficiency and generalization by transferring knowledge from complex, pre-trained models (teachers) to simpler models (students), often using meta-distillation framework or contrastive learning to prevent overfitting on limited data.
		
		Metric-based FSFGIC methods learn an embedding function, after which classification relies on a predefined or learned similarity measure in the feature space. Research in this area has advanced along several directions:$(1)$ Feature matching, e.g., framing image matching as an optimal transport problem~\cite{zhang2020deepemd} or reconstructing features via ridge regression~\cite{wertheimer2021few}, later extended to bidirectional reconstruction~\cite{wu2024bi1};$(2)$  Feature refinement and alignment, such as cross-layer and cross-sample optimization~\cite{ma2024cross}, progressive feature refinement~\cite{ma2025few}, and deformable convolution-based spatial–channel alignment~\cite{Huang2024FeatureAlignment}; and $(3)$ Attention mechanisms, including channel-spatial cross-attention~\cite{yang2024channel} and dual-attention frameworks~\cite{lee2024task}, which enhance discrimination by focusing on informative regions or channels.
		\subsection{Spatial-Frequency Fusion Methods}
		Recent research demonstrates that integrating spatial and frequency information significantly enhances model robustness in FSFGIC. FGFL~\cite{Cheng2023FGFL} analyzes the roles of different frequency bands and employs frequency-guided training to improve feature discrimination. MEFP~\cite{Zhou2024MEFP} decomposes images into distinct low-frequency (capturing global structures) and high-frequency (encoding fine details) branches. These are then aligned with spatial features via reconstruction to create more robust representations. FAP~\cite{Zhang2024FAP} embeds frequency cues into prompts that interact with spatial tokens, enhancing the model's transferability across different tasks or domains. Wavelet-MSFN~\cite{Wang2025WaveletMSFN} uses wavelet decomposition for multi-scale feature extraction, while SFIN-DPL~\cite{Liu2024SFIN_DPL} unifies spatial-frequency fusion with dual prompt learning for enhanced performance.  
		
		Based on our investigation, a research gap exists in leveraging phase information for FSFGIC. To address this, we propose a novel a phase information-based spatial and frequency information adaptively fusion network (PSF-Net). This approach significantly enhances the learning of robust and highly discriminative representations under strict few-shot fine-grained constraints.
		\section{Methodology}
		\subsection{Problem Defination}
		
		Formally, given a dataset $\mathcal{D}$ of $\mathcal{L}$ fine-grained categories, it is partitioned into three disjoint subsets: a training set $\mathcal{D}_{\text{train}}$, a validation set $\mathcal{D}_{\text{val}}$, and a test set $\mathcal{D}_{\text{test}}$, with $\mathcal{D}_{\text{train}}\cap\mathcal{D}_{\text{val}}\cap\mathcal{D}_{\text{test}}=\emptyset$ and $\mathcal{D}_{\text{train}}\cup\mathcal{D}_{\text{val}}\cup\mathcal{D}_{\text{test}}=\mathcal{D}$. The test set contains categories unseen during training. In each FSFGIC episode, a support set $\mathcal{S}$ and a query set $\mathcal{Q}$ are sampled. The support set contains $\mathcal{L}$ distinct classes with $\mathcal{K}$ labeled samples per class, while the query set consists of unlabeled samples from the same $\mathcal{L}$ classes. The objective is to learn an embedding function $f_{\theta}(\cdot)$ that maps an input image $X$ to a discriminative representation $z=f_{\theta}(X)$, enabling accurate classification of each query sample into its corresponding support class. This framework defines a standard $\mathcal{L}$-way $\mathcal{K}$-shot fine-grained classification task.
		\subsection{Overall Framework}
\begin{figure*}[htbp]
			\centering
			\vspace{-5pt}  
			\includegraphics[width=.99\linewidth]{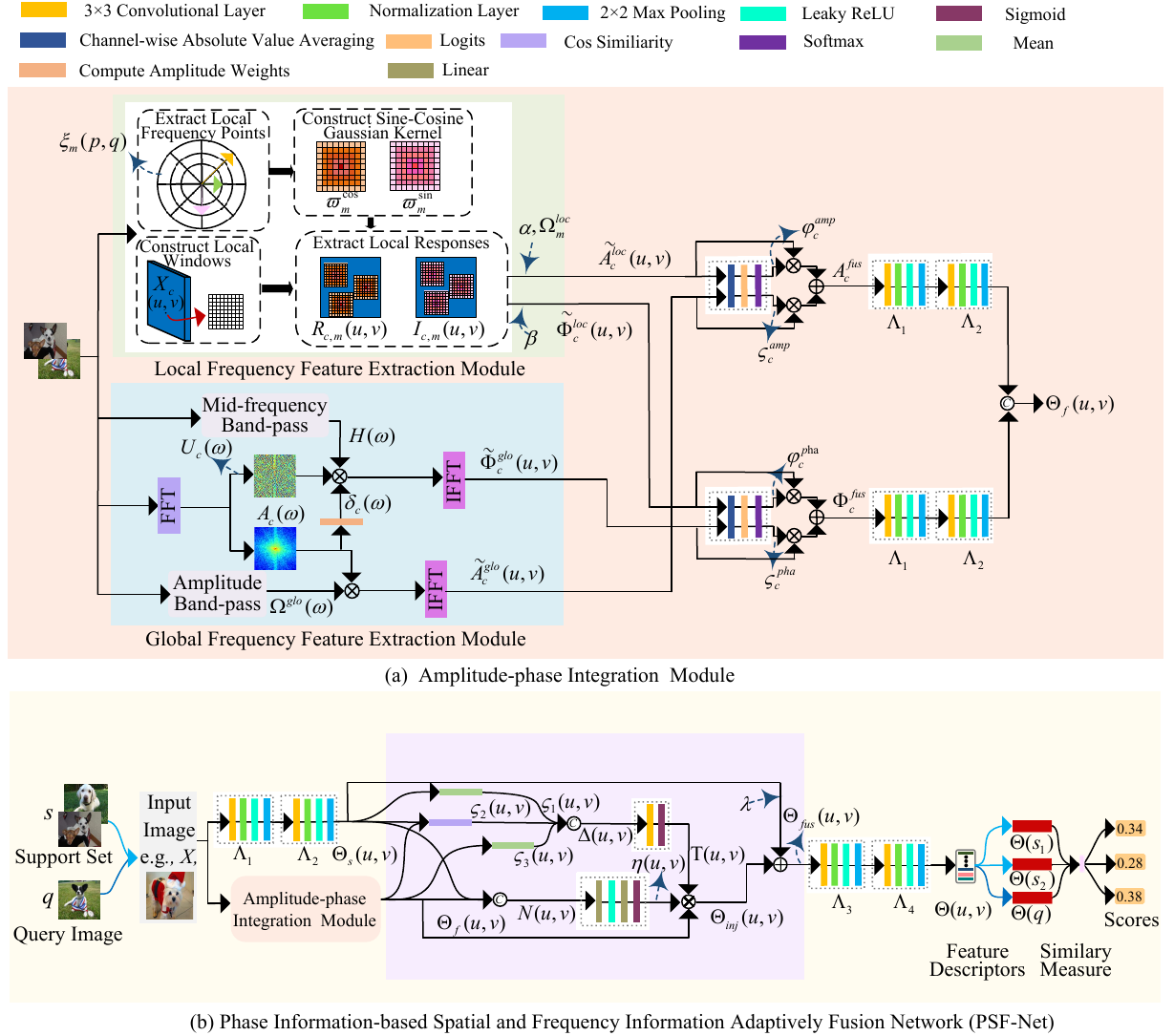}
			\setlength{\abovecaptionskip}{1pt}  
			\caption{The pipeline of the proposed PSF-Net for a 5-way 1-shot FSFGIC task based on the Conv-4 backbone.}
			\vspace{-5pt}  
			\label{fig2}
		\end{figure*}

		As shown in Fig.~\ref{fig2}(b), the designed PSF-Net contains three modules: a novel plug-and-play amplitude-phase integration (API) module, a spatial-frequency fusion module, and a similarity measurement module. 
		Following common FSFGIC practice (e.g., FRN~\cite{Wertheimer2021}, and
		SUITED~\cite{ma2025few}), we select Conv-4~\cite{snell20177} and ResNet12~\cite{lee2019meta} as backbones. We apply the phase-based spatial and frequency information adaptively for improving feature representations in fusion to the Conv-4 and the ResNet12 respectively. Then the similarity module is employed for measuring the distances between support and query samples.
		\subsection{Amplitude-Phase Integration (API) Module}
		This section introduces novel modules for extracting both local and global frequency features.
		
		(1) Local Frequency Feature Extraction Module: 
		Given an image $X(u,v) \in\mathbb{R}^{C \times H \times W}$,
		where $\mathbb{R}$ denotes the real space,
		$(u,v)$ denote the point location in the image,
		$C$ is the channel number, and $H$ and $W$ are the image height and width respectively.
		Local frequency magnitude and phase information of each channel $c \in \{1, \dots, C\}$ of the image is obtained as follows:
		\begin{equation}
			\begin{split}
				&{\xi}_m(p,q)={(r_i \cos \theta_j,r_i \sin \theta_j)},~\theta_j = \frac{2\pi j}{O}\\
				&\varpi_{m}^{\cos}(p,q) = \exp\Big(-\frac{p^2+q^2}{2\iota^2}\Big) \cos\big(2\pi \xi_{m}^\top (p,q)\big), \\
				&\varpi_{m}^{\sin}(p,q) = \exp\Big(-\frac{p^2+q^2}{2\iota^2}\Big) \sin\big(2\pi \xi_{m}^\top (p,q)\big),\\
				&R_{c,m}(u,v) =X_c(u,v) * \varpi_{m}^{\cos}(p,q), \\
				&I_{c,m}(u,v) = X_c(u,v) *  \varpi_{m}^{\sin}(p,q),\\
				&i\in\{1,\dots,R\},~j\in\{1,\dots,O\},~m\in\{1,\dots,R\times O\},
			\end{split}
		\end{equation}
		where ${\xi}_m(p,q)$ represents the local frequency points, $r_i$$\in$$\{0.30, 0.22, 0.16\}$ denotes the normalized radius of the $i$-th frequency scale (here $R$ is set to 3), $\theta_j$ $(j = 0,1,\dots,7)$ denotes the uniformly sampled orientation angles for each scale (here $O$ is set to 8), $\iota$ is adaptively optimized during training to adjust the effective receptive field of the kernel, $(p,q)$$\in$$[-4, 4]^2$ denotes a local $9$$\times$$9$ window, $\varpi_{m}^{\cos}(p,q)$ and $\varpi_{m}^{\sin}(p,q)$ are the Gaussian frequency kernels, $*$ denotes 2D convolution, and $R_{c,m}(u,v)$ and $I_{c,m}(u,v)$ denote the real and imaginary parts of the response of channel $c$ at spatial location $(u,v)$ to the $m$-th sampled frequency. Then, the complex response is converted into polar form for obtaining the local amplitude and the unit-phase representation.
		\begin{equation}
			\vspace{-3pt} 
			\begin{split}
				&A_{c,m}^{\mathrm{loc}}(u,v) = \sqrt{R_{c,m}^2(u,v) + I_{c,m}^2(u,v  )}, \\
				&{\phi}_{c,m}^{\mathrm{loc}}(u,v)  = \frac{1}{A_{c,m}^{\mathrm{loc}}}[u,v][R_{c,m}(u,v),I_{c,m}(u,v)]^\top,\\
				&\Omega_{m}^{\mathrm{loc}}(\xi) =
				\sigma\!\Big(\frac{\|\xi - \xi_{m}\|_2 - \kappa_{m}^{\mathrm{lo}}}{\tau_{m}}\Big)\sigma\!\Big(\frac{\kappa_{m}^{\mathrm{hi}} - \|\xi - \xi_{m}\|_2}{\tau_{m}}\Big),\\
				&\tilde{A}_c^{\mathrm{loc}}(u,v) = \sum_{m=1}^M\alpha\Omega_{m}^{\mathrm{loc}}A_{c,m}^{\mathrm{loc}}(u,v), \\
				&\tilde{{\Phi}}_c^{\mathrm{loc}}(u,v) = \sum_{m=1}^M \beta{\phi}_{c,m}^{\mathrm{loc}}(u,v),
			\end{split}
			\vspace{-3pt} 
		\end{equation}
		where $A_{c,m}^{\mathrm{loc}}(u,v)$ denotes the local magnitude response of the $c$-th channel of the $m$-th frequency component which reflects the activation strength of that frequency in the local region; where $\kappa_{m}^{\mathrm{lo}}$ and $\kappa_{m}^{\mathrm{hi}}$ are learnable cutoff radii, $\tau_{m}$ is a learnable parameter which is used to control the smoothness of the band, and $\xi$ is drawn from the predefined set of local frequency sampling points, $\sigma(\cdot)$ denotes the sigmoid function, ${\phi}_{c,m}^{\mathrm{loc}}(u,v)$ represents the corresponding unit-phase vector, encoding the directional information of the complex response and preserving the structural characteristics of the frequency component, $\alpha$ and $\beta$ are the adaptive weights for amplitude and phase respectively, $\Omega_{m}^{\mathrm{loc}}$ is the local amplitude component after band selection, and $\tilde{A}_c^{\mathrm{loc}}(u,v)$ and $\tilde{\Phi}_c^{\mathrm{loc}}(u,v)$ denote the final local amplitude and phase feature descriptors respectively. The obtained local amplitude and phase feature descriptors have the capability to enhance features relevant to fine texture regions while retaining subtle discriminative details and minor shape variation.
		(2) Global Frequency Feature Extraction Module:
		A global feature extraction module is presented for capturing the overall geometric structure and frequency information which aims to enhance the overall discriminative power of the representation. 
		
		The input image $X(u,v)$ is subjected to a channel-by-channel 2D real-valued fast Fourier transform (rFFT2) for obtaining the amplitude spectrum and phase spectrum information
		\begin{equation}
			\begin{aligned}
				&F_c(\boldsymbol{\omega})\! =\! \mathrm{rFFT2}(X_c)\triangleq A_c(\boldsymbol{\omega})\text{exp} \left(j\Phi_c(\boldsymbol{\omega})\right)\in\mathbb{Z}^{H \times W_r},\\
				&A_c(\boldsymbol{\omega}) = \lvert F_c(\boldsymbol{\omega}) \rvert,~U_c (\boldsymbol{\omega}) = \text{exp}\left(j\Phi_c(\boldsymbol{\omega})\right),
			\end{aligned}
		\end{equation}
		where $\mathbb{Z}$ denotes the complex space, $c$ is in $\{1,\dots,C\}$, $\boldsymbol{\omega}$$=$$(\omega_u,\omega_v)$ denote frequency coordinates, $W_r$ equals $W/2+1$, and $A_c(\boldsymbol{\omega})$ and $U_c(\boldsymbol{\omega})$ represent the amplitude and phase feature respectively.
		
		Motivated by~\cite{urban2011medium} that mid-frequency phase strikes a balance between spatial resolution and noise robustness, showing optimal performance for salient object localization, a mid-frequency band-pass $\mathcal{H}(\boldsymbol{\omega})$ is employed for selectively retaining components of the global phase spectrum within this frequency range, enabling more accurate localization of discriminative regions in the sample as follows:
		\begin{equation}
			\begin{aligned}
				&\bar{A}_c(\boldsymbol{\omega}) = \frac{1}{H W_r} \sum_{\boldsymbol{\omega}} A_c(\boldsymbol{\omega}),~\delta_c(\boldsymbol{\omega})=\left( \frac{A_c(\boldsymbol{\omega})}{\bar{A}_c(\boldsymbol{\omega})} \right)^l, \\
				&\mathcal{H}(\boldsymbol{\omega}) =
				\begin{cases}
					1, & 0.05 \le \ \rho(\boldsymbol{\omega}) \le 0.35, \\
					0, & \text{otherwise},
				\end{cases} \\
				&\tilde{\Phi}_c(\boldsymbol{\omega}) =U_c(\boldsymbol{\omega})\delta_c(\boldsymbol{\omega})\mathcal{H}(\boldsymbol{\omega}),\\
				&\Omega^{\mathrm{glo}}(\boldsymbol{\omega}) =
				\sigma\!\left(
				\frac{\rho(\boldsymbol{\omega}) - \psi^{\mathrm{lo}}}{\tau}
				\right)
				\sigma\!\left(
				\frac{\psi^{\mathrm{hi}} - \rho(\boldsymbol{\omega})}{\tau}
				\right),\\
				&\tilde{A}_c(\boldsymbol{\omega})=\Omega^{\mathrm{glo}}A_c(\boldsymbol{\omega}),\\
			\end{aligned}
		\end{equation}
		where $\bar{A}_c(\boldsymbol{\omega})$ denotes the mean amplitude of the $c$-th channel, $l$$\in$$[0,1]$ is a learnable parameter which is employed for controling the strength of the amplitude modulation, $\psi^{\mathrm{lo}}$ and $\psi^{\mathrm{hi}}$ are learnable cutoff radii, $\tau$ controls the transition smoothness, and $\rho(\boldsymbol{\omega})$$=$$\sqrt{\boldsymbol{\omega}_u^2 + \boldsymbol{\omega}_v^2}$ represents the radial frequency. With this strategy, the phase feature has the capability to precisely encode the global geometric layout and overall shape of the target. Furthermore, the processed global frequency information is transformed back to the spatial domain via the inverse Fourier transform,  yielding features that can be directly used for subsequent local-global feature fusion
		\begin{equation}
			\begin{split}
				\vspace{-5pt} 
				\tilde{A}_c^{\mathrm{glo}}(u, v) =
				\mathrm{irFFT2}\!\left(\tilde{A}_c(\boldsymbol{\omega})\right),\\
				\tilde{\Phi}_c^{\mathrm{glo}}(u, v) =
				\mathrm{irFFT2}\!\left(
				\tilde{\Phi}_c(\boldsymbol{\omega})
				\right).
			\end{split}
		\end{equation}
		
		(3) Local-global Frequency Information Fusion:
		A channel-energy-driven local-global adaptive frequency information fusion is designed for fully leveraging phase information under limited training samples. For each channel, the average absolute energies of the local and global feature information in the spatial domain are computed as:
		\begin{equation}
			\begin{split}
				&\zeta_c^{\mathrm{amp}}
				= \mathrm{Softmax}\!\left(
				\gamma_1 + \log\!\left(
				\frac{1}{HW}\!
				\sum_{u,v} \big|\tilde{A}_c^{\mathrm{loc}}(u,v)\big|
				\right)
				\right),\\
				&\zeta_c^{\mathrm{pha}} = \text{Softmax}\!\left( \gamma_2 + \log\!\left(\!\frac{1}{HW}\!\sum_{u,v}\!\big| \tilde{\Phi}_c^{\mathrm{loc}}(u, v) \big|\right)
				\right),\\
				&\varphi_c^{\mathrm{amp}} = \text{Softmax}\!\left( \gamma_3 + \log\!\left(\!\frac{1}{HW}\!\sum_{u,v}\!\big| \tilde A_c^{\mathrm{glo}}(u, v) \big|\right)
				\right),\\
				& \varphi_c^{\mathrm{pha}} = \text{Softmax}\!\left( \gamma_4 + \log\!\left(\!\frac{1}{HW}\!\sum_{u,v}\!\big| \tilde{\Phi}_c^{\mathrm{glo}}(u, v) \big|\right)
				\right),
			\end{split}
		\end{equation}
		where $\gamma_1$, $\gamma_2$, $\gamma_3$, and $\gamma_4$ are learnable bias terms, $\log(\cdot)$ denotes a logarithmic compression operator to stabilize the dynamic range of frequency energy and prevent dominance of overwhelmingly strong low-frequency responses, $\zeta_c^{\text{amp}}$ and $\zeta_c^{\text{pha}}$ represent local amplitude and phase weight parameters respectively, and $\varphi_c^{\text{amp}}$ and $\varphi_c^{\text{pha}}$ are global amplitude and phase weight parameters respectively. The local and global amplitude and phase descriptors for each channel are then fused respectively for obtaining the frequency-enhanced feature descriptors $\Theta_f(u,v)$ as follows:
		\begin{equation}
			\label{eq2}
			\begin{split}    
				A_c^{\mathrm{fus}}(u,v) = \zeta_c^{\mathrm{amp}} \, \tilde{A}_c^{\mathrm{loc}}(u,v) + \varphi_c^{\mathrm{amp}} \, \tilde{A}_c^{\mathrm{glo}}(u,v),\\
				{\Phi}_c^{\mathrm{fus}}(u,v) = \zeta_c^{\mathrm{pha}} \, \tilde{\Phi}_c^{\mathrm{loc}}(u,v) + \varphi_c^{\mathrm{pha}} \, \tilde{\Phi}_c^{\mathrm{glo}}(u,v).
			\end{split}
		\end{equation} 
		After feature descriptors $A_c^{\mathrm{fus}}$ and $\Phi_c^{\mathrm{fus}}$ pass through the convolutional block $\Lambda_1$ and $\Lambda_2$ respectively, the obtained $\hat{A}_c^{\mathrm{fus}}$ and $\hat{\Phi}_c^{\mathrm{fus}}$ are concatenated for obtaining feature descriptors $\Theta_f(u,v)$ which has the capability to capture global structural information while preserving local detail resolution.
		\subsection{Spatial-Frequency Fusion Module}
		Phase information, which mainly captures structural variations in mid-to-high frequencies, tends to be attenuated by downsampling and deep semantic aggregation. To preserve these cues, we inject frequency-enhanced features $\Theta_f(u,v)$ at shallow stages, using the output $\Theta_s(u,v)$ of the first two convolutional blocks as spatial structural features. As shown in Fig.~\ref{fig2}(a), this allows geometric guidance to continuously steer the backbone from early extraction, enhancing robustness to device variation in few-shot fine-grained recognition.
		
		The fused spatial and frequency feature descriptors $\Theta_{\mathrm{fus}}(u,v)$ are obtained as follows:
		\begin{equation}
			\begin{split}
				&N(u,v)=\mathrm{Concat}\big({\Theta}_s(u,v),{\Theta}_f(u,v)\big),\\ 
				&\eta(u,v) = \sigma\!\big(\mathrm{MLP}(N(u,v))\big), \\
				&\varsigma_1(u,v)=\text{cos}({\Theta}_s(u,v),{\Theta}_f(u,v)),\\
				&\varsigma_2(u,v)=\mathrm{mean}({\Theta}_s(u,v)),~\varsigma_3(u,v)=\mathrm{mean}({\Theta}_f(u,v)),\\
				&\Delta(u,v)=\mathrm{Concat}\big(\varsigma_1(u,v),\varsigma_2(u,v),\varsigma_3(u,v)\big),\\
				&\mathcal{T}(u,v) = \sigma(\mathrm{Conv}_{3\times3}(\Delta(u,v))), \\
				&\Theta_{\mathrm{inj}}(u,v) = {\Theta}_f(u,v) \odot \eta(u,v)\odot \mathcal{T}(u,v),\\
				&\Theta_{\mathrm{fus}}(u,v) = \lambda{\Theta}_s(u,v)+\Theta_{\mathrm{inj}}(u,v),
			\end{split}
		\end{equation}
		where $\cos(\cdot,\cdot)$ represents cosine similarity, $\mathrm{MLP}$ is the multiple layer perceptron operation, $\mathrm{mean}(\cdot)$ is the channel-wise average response at each spatial location, $\odot$ is the dot product operator, and $\lambda$ is a learnable weight factor. The fused feature descriptors $\Theta_{\mathrm{fus}}(u,v)$ are sent into the following two convolutional blocks for obtaining the feature representations $\Theta(u,v)$.
		\subsection{Similarity Measure Module}
		The feature descriptor $\Theta(X)$$\in$$\mathbb{R}^{h\times w\times d}$ obtained from the designed PSF-Net is reshaped to tokens $\bar{\Theta}(X)$$\in$$\mathbb{R}^{t\times d}$ with $t$$=$$hw$, where $d$ is the number of channels, $h$ and $w$ denote the height and the width of feature maps. For an $\mathcal{L}$-way $\mathcal{K}$-shot episode, let $n$$\in$$\{1,\dots,\mathcal{L}\}$ index the sampled classes,
		$\{s_{n,k}\}_{k=1}^{\mathcal{K}}$ denote the $\mathcal{K}$ support images of class $n$, and $q$ denote a query image.
		We obtain token descriptors $\bar{Q}$$=$$\bar{\Theta}(q)\in\mathbb{R}^{t\times d}$ and
		\begin{equation}
			\bar{S}_n=\mathrm{Concat}_{k=1}^{\mathcal{K}}\bar{\Theta}(s_{n,k})\in\mathbb{R}^{(\mathcal{K}t)\times d}.
		\end{equation}
		Following~\cite{wu2024bi1}, we use a lightweight bidirectional cross-attention head, where
		$\mathrm{Attn}(Q,K,V) = \mathrm{Softmax}\!\left(\frac{QK^\top}{\sqrt{d}}\right)V$,
		with $Q$, $K$, and $V$ linearly projected from tokens (see~\cite{wu2024bi1} for details).  
		For each class $n$, the bidirectional reconstructions are:
		\begin{equation}	\hat{Q}_n=\mathrm{Attn}(\bar{Q}^{Q},\bar{S}_n^{K},\bar{S}_n^{V}),~\hat{S}_n=\mathrm{Attn}(\bar{S}_n^{Q},\bar{Q}^{K},\bar{Q}^{V}),
		\end{equation}
		and we compute
		\begin{equation}
			\begin{split}
				&d_n=\lambda_1\|\bar{Q}^{V}-\hat{Q}_n\|^2+\lambda_2\|\bar{S}_n^{V}-\hat{S}_n\|^2,\\
				&P(y=n\mid q)=\frac{\exp(-d_n)}{\sum_{n=1}^{\mathcal{L}}\exp(-d_n)}.
			\end{split}
		\end{equation}
		The whole model is trained end-to-end with episode-wise cross-entropy on $P(y\mid q)$.
		\begin{table*}[t]
			\caption{Comparison results of different methods for 5-way tasks on the CUB-200-2011, Stanford Dogs, and Stanford Cars datasets with two different backbones. The best performance is indicated in bold.}
			
			\label{t4}
			\centering
			\footnotesize
			\setlength{\tabcolsep}{5pt}
			\renewcommand{\arraystretch}{1.2}
			\begin{tabular}{@{}c|c|cc|cc|cc@{}}
				\hline
				\multirow{2}{*}{\centering Backbone} &
				\multirow{2}{*}{\centering Method} &
				\multicolumn{2}{c|}{CUB-200-2011} &
				\multicolumn{2}{c|}{Stanford Dogs} &
				\multicolumn{2}{c}{Stanford Cars} \\
				\cline{3-8}
				& & 1-shot & 5-shot & 1-shot & 5-shot & 1-shot & 5-shot \\
				\hline
				
				\multirow{10}{*}{\centering Conv-4}
				& ProtoNet~\cite{snell20177} & 64.82 & 85.74 & 46.66 & 70.77 & 50.88 & 74.89 \\
				
				
				& FRN~\cite{Wertheimer2021} & 74.90 & 89.39 & 60.41 & 79.26 & 67.48 & 87.97 \\
				
				
				& DAN~\cite{Shulin2022} & 72.89 & 86.60 & 59.81 & 77.19 & 70.21 & 85.55 \\
				
				& DeepEMD~\cite{zhang2020deepemd} & 64.08 & 80.55 & 46.73 & 65.74 & 61.63 & 72.95 \\
				
				
				& FRN+CSCAM~\cite{yang2024channel} & 77.68 & 89.88 & - & - & 71.44 & 86.44 \\
				
				
				& SUITED~\cite{ma2025few} & 79.73 & 90.05 & 68.67 & 82.24 & 82.21 & 92.39 \\
				
				& BDFRNet$\dagger$~\cite{wu2024bi1} & 76.39 & 90.61 & 64.66 & 81.27 & 75.33 & 90.91 \\
				
				& C2-Net$\dagger$~\cite{ma2024cross} & 78.63 & 89.48 & 69.81 & 84.39 & 79.52±0.45 & 91.15 \\
				
				\cline{2-8}
				
				& Ours & \textbf{80.80} & \textbf{92.70} & \textbf{70.40} & \textbf{85.30} & \textbf{82.28} & \textbf{94.58} \\
				
				& Ours-Snapshot & \textbf{83.80} & \textbf{94.05} & \textbf{72.80} & \textbf{86.90} & \textbf{84.40} & \textbf{95.20} \\
				
				\hline
				
				\multirow{9}{*}{\centering ResNet12}
				& FRN~\cite{Wertheimer2021} & 82.86 & 92.41 & 76.76 & 88.74 & 86.90 & 95.69 \\
				
				& DeepEMD~\cite{zhang2020deepemd} & 75.59 & 88.23 & 70.38 & 85.24 & 80.62±0.26 & 92.633 \\
				
				& FRN+CSCAM~\cite{yang2024channel} & 84.00 & 93.52 & - & - & 86.24 & 95.55 \\
				
				& TDM+CSCAM~\cite{yang2024channel} & 83.34 & 92.98 & - & - & 86.86 & 95.63 \\
				
				& SUITED$\dagger$~\cite{ma2025few} & 83.11 & 92.01 & 76.55 & 88.867 & 89.90 & 96.53 \\
				
				& BDFRNet$\dagger$~\cite{wu2024bi1} & 82.03 & 92.78 & 77.40 & 88.41 & 90.28 & 97.26 \\
				
				& C2-Net$\dagger$~\cite{ma2024cross} & 83.65 & 92.57 & 77.72 & 89.59 & 86.48 & 94.07 \\
				
				\cline{2-8}
				
				& Ours & \textbf{84.05} & \textbf{94.01} & \textbf{77.94} & \textbf{89.87} & \textbf{90.50} & \textbf{97.67} \\
				
				& Ours-Snapshot & \textbf{85.22} & \textbf{94.82} & \textbf{80.70} & \textbf{91.45} & \textbf{91.60} & \textbf{97.87} \\
				
				\hline
			\end{tabular}
		\end{table*}
		
		\section{Experiments}
		
		\subsection{Datasets}
		The proposed PSF-Net is evaluated on five standard FSFGIC benchmarks: CUB-200-2011~\cite{wah2011caltech}, Stanford Dogs~\cite{khosla2011novel}, Stanford Cars~\cite{krause20133d}, meta-iNat~\cite{wertheimer2019few}, and tiered meta-iNat~\cite{wertheimer2019few}. CUB-200-2011 contains 200 bird species with 11,788 images, representing a classic benchmark for fine-grained recognition. Stanford Dogs comprises 20,580 images covering 120 dog breeds, Stanford Cars includes 16,185 images from 196 car categories, meta-iNat provides 1,135 fine-grained wildlife categories, while tiered meta-iNat introduces a more challenging evaluation setting by increasing the domain gap between base and novel classes. All experiments adhere to the standard dataset splits (summarized in Table~\ref{tab2}) to ensure a fair comparisons with prior works.
		\begin{table}[t]
			\caption{Class split of the five datasets. $N_{\text{train}}$, $N_{\text{val}}$, and $N_{\text{test}}$ denote the numbers of classes in the training, validation, and test sets, respectively.}
			\label{tab2}
			\centering
			\renewcommand{\arraystretch}{1.2}
			\begin{tabular}{lccc}
				\hline
				Dataset & $N_{\text{train}}$ & $N_{\text{val}}$ & $N_{\text{test}}$ \\
				\hline
				CUB-200-2011 & 100 & 50 & 50 \\
				Stanford Dogs & 70 & 20 & 30 \\
				Stanford Cars & 130 & 17 & 49 \\
				meta-iNat & 908 & -- & 227 \\
				tiered meta-iNat & 781 & -- & 354 \\
				\hline
			\end{tabular}
		\end{table}
		
		\subsection{Experimental Setup}
		Experiments were performed on the five fine-grained datasets under standard 5-way 1-shot and 5-shot FSFGIC protocols. The proposed PSF-Net was trained end-to-end from scratch, with unique parameters in each convolutional block. Each input image was first resized to 92$\times$92 pixels and then randomly cropped to 84$\times$84 during training.
		
		We adopted Conv-4~\cite{snell20177} and ResNet-12~\cite{lee2019meta} backbones, training all models for 1,200 epochs using SGD with Nesterov momentum (0.9) and a weight decay of \(5 \times 10^{-4}\). The initial learning rate was set to 0.1 and reduced by a factor of 10 for every 400 epochs. To manage memory usage, the Conv-4 models were trained with 30-way 5-shot episodes, while the ResNet-12 models used 15-way 5-shot episodes. All models were evaluated on the standard 5-way 1-shot and 5-shot tasks. Validation was carried out for every 20 epochs, and the model with the best validation performance was selected for final testing. Reported accuracy is the mean over 10,000 randomly sampled test episodes for both 5-way 1-shot and 5-shot settings.
		\begin{table}[t]
			\caption{Comparison results of different methods on the meta-iNat and tiered meta-iNat datasets in the 5-way setting under the Conv-4 backbone. The best performance is indicated in bold.}
			\label{t3}
			\centering
			\scriptsize
			\setlength{\tabcolsep}{15pt}
			\renewcommand{\arraystretch}{1.2}
			
			\begin{tabular}{l|cc|cc}
				\hline
				\multirow{2}{*}{\centering Method} & \multicolumn{2}{c|}{meta-iNat} & \multicolumn{2}{c}{tiered meta-iNat} \\
				\cline{2-5}
				& 1-shot & 5-shot & 1-shot & 5-shot \\
				\hline
				
				ProtoNet~\cite{snell20177} & 53.78 & 73.80 & 35.47 & 54.85 \\
				FRN~\cite{Wertheimer2021} & 61.98 & 80.04 & 43.95 & 63.45 \\
				DeepEMD~\cite{zhang2020deepemd} & 54.48 & 68.36 & 36.05 & 48.55 \\
				C2-Net~\cite{ma2024cross} & 71.47 & 85.47 & 49.04 & 67.25 \\
				TDM+IAM~\cite{lee2024task} & 65.95 & 83.30 & 46.45 & 66.55 \\
				BDFRNet~\cite{wu2024bi1} & 66.07 & 83.30 & 46.64 & 66.46 \\
				SUITED~\cite{ma2025few} & 74.72 & 87.44 & 51.70 & 70.43 \\
				\hline
				
				Ours & \textbf{74.85} & \textbf{89.00} & \textbf{52.43} & \textbf{72.77} \\
				Ours-Snapshot & \textbf{76.29} & \textbf{89.80} & \textbf{55.51} & \textbf{75.89} \\
				
				\hline
			\end{tabular}
			\vspace*{-10pt} 
		\end{table}
		For snapshot training, we adopt a cosine annealing learning rate schedule, saving a model snapshot at each minimum (every 240 epochs) to create a 5-model ensemble for validation-based selection. During inference, the final prediction is obtained by averaging the output probabilities of all snapshots, and the mean accuracy and variance on the test set are reported.
		\subsection{Performance Comparison}
\begin{figure}[htbp]
			\centering
			\includegraphics[width=0.95\linewidth]{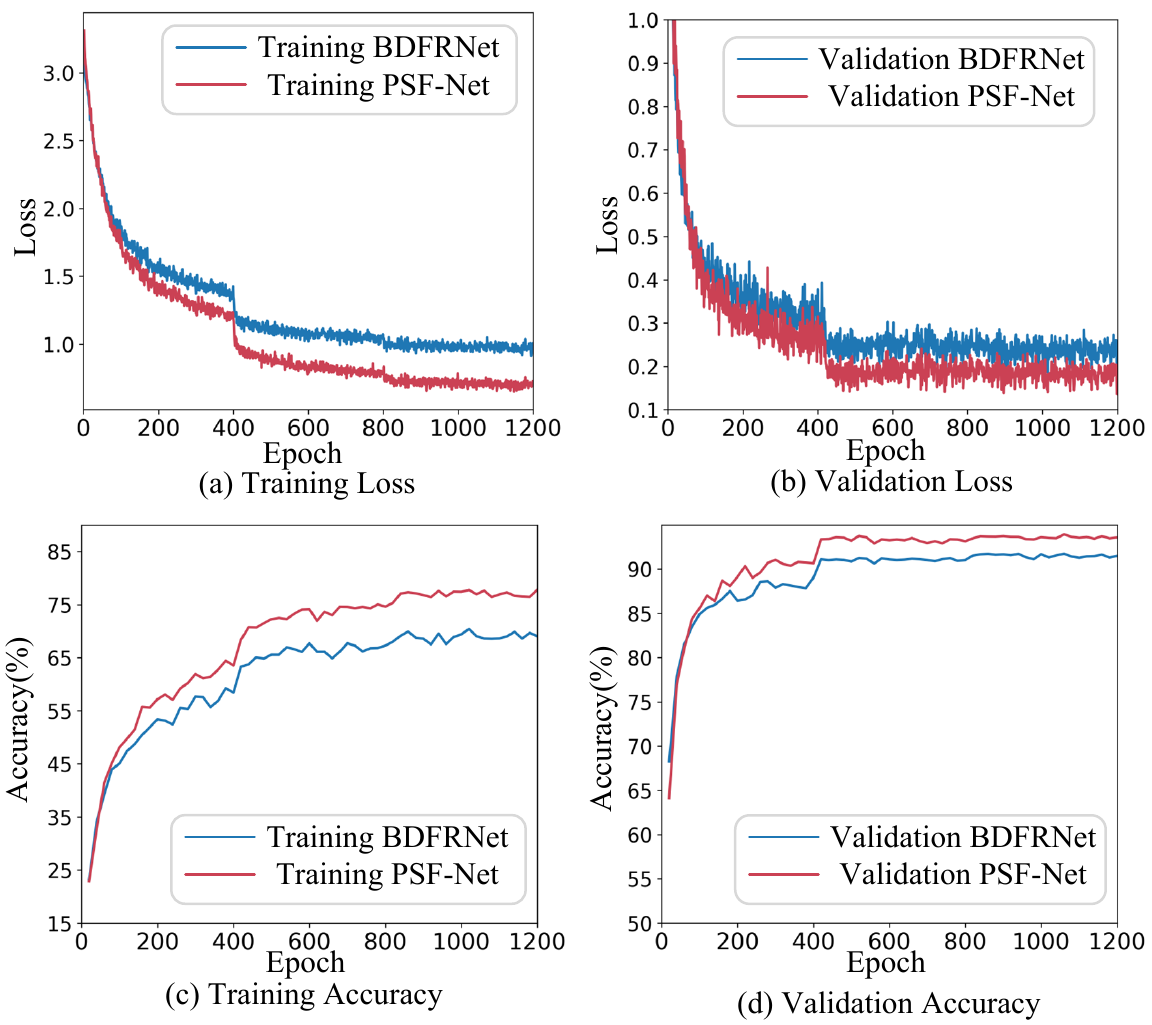}
			\caption{Examples of the loss and accuracy curves of BDFRNet and the proposed method for the 5-way 1-shot FSFGIC task on the CUB-200-2011 dataset.}
			\label{fig:7}
		\end{figure}

		A comparative analysis between the proposed PSF-Net and Nine state-of-the-art methods is illustrated. As summarized in Table~\ref{t4} and Table~\ref{t3}, our method achieves the best performance on five evaluated datasets. For instance, on the CUB-200-2011 dataset using a Conv-4 backbone, PSF-Net achieves accuracies of 83.80$\%$ (5-way 1-shot) and 94.05$\%$ (5-way 5-shot), outperforming the current best alternative, C2-Net (78.63$\%$ and 89.48$\%$, respectively)~\cite{ma2024cross}. These results validate the efficacy of our designed phase-sensitive spatial–frequency few-shot network.
		Taking the 5-way 1-shot task on the CUB-200-2011 dataset as an example, Fig.~\ref{fig:7} compares the training and validation performance of BDFRNet and the proposed PSF-Net. The accuracy curves (Fig.~\ref{fig:7}(a) and (b)) and loss curves (Fig.~\ref{fig:7}(c) and (d)) demonstrate that PSF-Net achieves consistently lower loss and higher accuracy than BDFRNet throughout both training and validation phases.
		\subsection{Ablation Study on Fusion Depth}
		To determine the most suitable position for integrating the proposed API module, we apply the API at different depth stages of the backbone. Each stage is designed as a convolutional block (comprising convolution, normalization, activation, and pooling), denoted as $\Lambda_i(i=1,2,3,4)$.We follow the same experimental setup as before, conducting experiments based on the Conv4 and CUB dataset. As shown in Table~\ref{tab:fusion_depth}, the best performance is
		obtained when fusion is applied at $\Lambda_2$. This observation aligns with the characteristics
		of few-shot fine-grained tasks: the most discriminative cues lie in subtle local structural
		regions. Phase information sensitively encodes these geometry patterns; however, deeper
		semantic abstraction weakens its expressiveness due to reduced spatial resolution. $\Lambda_2$
		provides an optimal trade-off, where part-level structure is preserved while shallow
		appearance noise is largely suppressed. Therefore, $\Lambda_2$ is adopted as the default setting
		in our main experiments.
		\begin{table}
			\caption{Performance of the API module with Conv-4 on CUB-200-2011 under different fusion depth settings (5-way).}
			\label{tab:fusion_depth}
			\renewcommand{\arraystretch}{0.95}
			\centering
			\begin{tabular}{@{}c c cc@{}}
				\toprule[1.pt]
				Fusion location & Retained information & 1-shot & 5-shot \\
				\midrule
				$\Lambda_1$ & high-resolution textures & 78.40±0.20 & 90.67±0.21 \\
				\textbf{$\Lambda_2$ (ours)} & local part structures & \textbf{80.80±0.19} & \textbf{92.71±0.10} \\
				$\Lambda_3$ & semantic parts & 80.08±0.20 & 91.71±0.09 \\
				$\Lambda_4$ & global semantics & 79.88±0.18 & 92.01±0.08 \\
				\bottomrule[1.pt]
			\end{tabular}
			\vspace{-10pt}
		\end{table}
		\subsection{Ablation Study for different frequency components}
		To further investigate the effectiveness of our proposed method, ablation experiments are conducted on the CUB-200-2011, Stanford Dogs, and Stanford Cars datasets as follows.
		
		We compare three configurations to investigate the contribution of different frequency components:
		(i) a spatial-only model (PSF-Net$_{S}$); (ii) a spatial–frequency model without phase information (PSF-Net$_{A}$),which uses only amplitude; (iii)  a spatial–frequency modelwithout amplitude information (PSF-Net$_{P}$), and (iv) a full spatial–frequency model with both amplitude and phase (PSF-Net).As shown in Table~\ref{t5}, the complete model that incorporates phase achieves the best performance across all datasets for both Conv-4 and ResNet-12 backbones, indicating that phase-aware frequency cues provide additional complementary information beyond spatial and amplitude features.										
		\begin{figure*}[t]
			\setlength{\abovecaptionskip}{-1pt}
			\setlength{\belowcaptionskip}{-10pt}
			\centering
			\includegraphics[width=0.95\textwidth]{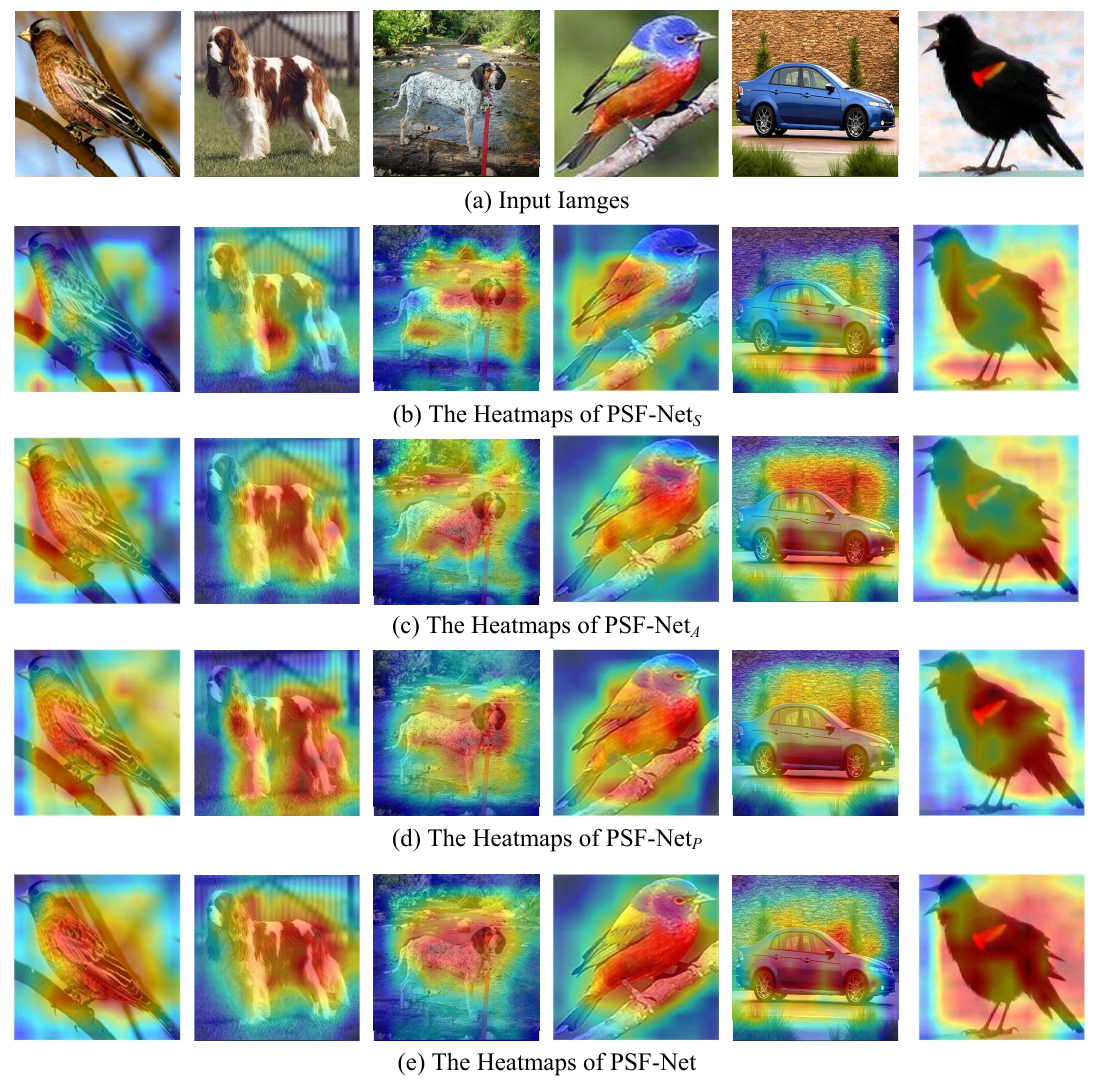}
			\caption{The heatmaps of six images visualized by different frequency components.}
			\label{fig:8}
		\end{figure*}
		\subsection{Ablation Study on Local Frequency Sampling Strategy}
		
		To evaluate the generality and optimality of the proposed local frequency sampling strategy, we conduct comprehensive ablation studies on CUB-200-2011 and Stanford Cars datasets under the 5-way 1-shot/5-shot settings with a Conv-4 backbone. Only the radius set and the number of orientations in the local frequency module are modified. The default configuration ($r=\{0.30, 0.22, 0.16\}$, $o=8$) is designed to capture discriminative mid-to -high frequency structural cues across multiple scales while maintaining sufficient orientation sensitivity to part-level geometric variations, without introducing excessive learnable parameters. Furthermore, phase-based representations are more responsive to subtle structural boundaries, which is essential in fine-grained recognition under limited data.
		\begin{table*}[t]
			\caption{Ablation study of local frequency sampling using the Conv-4 backbone.}
			\label{tab:ablation_lfs}
			\centering
			\scriptsize
			\setlength{\tabcolsep}{3pt}
			\renewcommand{\arraystretch}{1.2}
			\begin{tabular}{c c cc cc}
				\toprule[1pt]
				Model & Setting ($r$, $o$) & \multicolumn{2}{c}{CUB-200-2011} & \multicolumn{2}{c}{Stanford Cars} \\
				& & 1-shot & 5-shot & 1-shot & 5-shot \\
				\midrule
				$r_1$ & $\{0.30\},\,8$ &
				79.45$\pm$0.20 & 91.12$\pm$0.65 &
				80.90$\pm$0.19 & 93.10$\pm$0.08 \\
				
				$r_2$ & $\{0.30,0.22\},\,8$ &
				79.80$\pm$0.18 & 91.65$\pm$0.60 &
				81.35$\pm$0.17 & 93.55$\pm$0.07 \\
				
				$r_3$ & $\{0.30,0.22,0.16\},\,4$ &
				80.10$\pm$0.17 & 92.00$\pm$0.58 &
				81.70$\pm$0.16 & 94.00$\pm$0.06 \\
				
				Ours & $\{0.30,0.22,0.16\},\,8$ &
				\textbf{80.81$\pm$0.18} & \textbf{92.69$\pm$0.6} &
				\textbf{82.28$\pm$0.17} & \textbf{94.58$\pm$0.07} \\
				
				$r_4$ & $\{0.30,0.22,0.16,0.12\},\,8$ &
				80.70$\pm$0.18 & 92.60$\pm$0.60 &
				82.10$\pm$0.17 & 94.50$\pm$0.07 \\
				
				$r_5$ & $\{0.30,0.22,0.16\},\,12$ &
				80.75$\pm$0.18 & 92.65$\pm$0.60 &
				82.15$\pm$0.17 & 94.55$\pm$0.07 \\
				
				$r_6$ & $\{0.30,0.22,0.16\},\,16$ &
				80.78$\pm$0.18 & 92.67$\pm$0.60 &
				82.20$\pm$0.17 & 94.57$\pm$0.07 \\
				
				\bottomrule[1pt]
			\end{tabular}
		\end{table*}

		Results demonstrate that increasing the number of radii progressively improves performance, validating the effectiveness of multi-scale phase structural encoding in distinguishing subtle fine-grained differences. In contrast, insufficient orientation coverage degrades geometric sensitivity, while excessively dense radius or orientation settings introduce redundant high-frequency noise and increased overfitting risk.As shown in Table~\ref{t2},  the configuration of $r=\{0.30,0.22,0.16\}$ and $o=8$ achieves the best trade-off between representation capacity and generalization, proving the efficiency and robustness of our local frequency sampling strategy.

		\subsection{Grad-CAM Visualization}
		Meanwhile,to further illustrate the effectiveness of the proposed PSF-Net, we employ the Grad-CAM visualization technique
		using ResNet-12 to analyze model attention regions. It utilizes the six images shown in Fig.~\ref{fig:8}(a) for illustration. In Grad-CAM, regions with higher energies denote more discriminative parts of an image. Fig.~\ref{fig:8}(b), (c), (d) and (e) display the attention maps of these six images : (i) a spatial-only model (PSF-Net$_{S}$); (ii) a spatial–frequency model without phase information (PSF-Net$_{A}$),which uses only amplitude; (iii)  a spatial–frequency modelwithout amplitude information (PSF-Net$_{P}$), and (iv) a full spatial–frequency model with both amplitude and phase (PSF-Net). The proposed PSF-Net with spatial-frequency domain demonstrates a stronger ability to focus on the classification targets themselves.
		\begin{table*}[!t]
			\centering
			\caption{Performance across different frequency components on the proposed PSF-Net with Conv-4 and ResNet-12 using the CUB-200-2011, Stanford Dogs, and Stanford Cars datasets for 5-way tasks.}
			\renewcommand{\arraystretch}{1}
			{\begin{tabular}{c|c|cc|cc|cc}
					\toprule[1.pt]
					\multirow{2}{*}{Model} &  \multirow{2}{*}{Backbone}  & \multicolumn{2}{c|}{CUB-200-2011} & \multicolumn{2}{c|}{Stanford Dogs} &  \multicolumn{2}{c}{Stanford Cars}\\
					\cline{3-8}
					&  &  1-shot & 5-shot & 1-shot & 5-shot  & 1-shot & 5-shot \\
					\cline{1-8}
					
					PSF-Net$_{S}$  & Conv-4  &76.39 &90.61   &64.66 &81.27 &75.33 &90.91 \\
					PSF-Net$_{A}$  & Conv-4  & 80.02   & 92.28   & 66.86   & 82.57   & 78.27  & 92.78 \\
					PSF-Net$_{P}$  & Conv-4  & 79.52   & 92.00   & 69.05   & 84.27   & 79.83  & 93.54 \\
					PSF-Net  & Conv-4  & \textbf{80.90}     & \textbf{92.81}    & \textbf{70.50}   & \textbf{85.30}   & \textbf{82.28}  & \textbf{94.51} \\
					
					\hline
					PSF-Net$_{S}$   & ResNet-12  &82.03 &92.78    &77.40   &88.41 &90.28 &97.26 \\
					PSF-Net$_{A}$  & ResNet-12 & 83.95      & 93.87     & 77.67   & 85.45   & 90.48   & 97.53 \\
					PSF-Net$_{P}$  & ResNet-12 & 82.26      & 92.99     & 77.61   & 89.56   & 90.43   & 97.69 \\
					PSF-Net & ResNet-12   & \textbf{84.06}     & \textbf{94.02}    & \textbf{77.95}   & \textbf{89.86}   & \textbf{90.62}  & \textbf{97.73} \\			
					\bottomrule[1.pt]
			\end{tabular}}
			\label{t5}
		\end{table*}
		
		\section{Conclusion}
		This work focuses on the challenge of learning discriminative features for few-shot fine-grained classification when only limited samples are available. We introduce a spatial–frequency framework that incorporates explicit phase modeling at the early feature extraction stage and uses an energy-guided adaptive fusion module to enhance structure-sensitive cues. These designs enable the model to better exploit complementary spatial and frequency information and improve fine-grained recognition under limited-sample settings. The proposed framework is plug-and-play, and compatible with common backbone networks, achieving consistent gains across multiple benchmarks. Overall, our findings highlight the critical role of phase information and frequency-domain energy statistics in improving structural separability and offer a general paradigm for advancing few-shot visual recognition.

		
		\bibliographystyle{elsarticle-num}
		\bibliography{Ref}
		
	\end{document}